# CLAMP: Contrastive Learning with Adaptive Multi-loss and Progressive Fusion for Multimodal Aspect-Based Sentiment Analysis

Xiaoqiang He

[a] School of Information Engineering, Minzu University of China, Beijing, 100081, China

Abstract: Multimodal aspect-based sentiment analysis (MABSA) seeks to identify aspect terms within paired image–text data and determine their fine-grained sentiment polarities, representing a fundamental task for improving the effectiveness of applications such as product review systems and public opinion monitoring. Existing methods face challenges such as cross-modal alignment noise and insufficient consistency in fine-grained representations. While global modality alignment methods often overlook the connection between aspect terms and their corresponding local visual regions, bridging the representation gap between text and images remains a challenge. To address these limitations, this paper introduces an end-to-end Contrastive Learning framework with Adaptive Multi-loss and Progressive Attention Fusion (CLAMP). The framework is composed of three novel modules: Progressive Attention Fusion network, Multi-task Contrastive Learning, and Adaptive Multi-loss Aggregation. The Progressive Attention Fusion network enhances fine-grained alignment between textual features and image regions via hierarchical, multi-stage cross-modal interactions, effectively suppressing irrelevant visual noise. Secondly, multi-task contrastive learning combines global modal contrast and local granularity alignment to enhance cross-modal representation consistency. Adaptive Multi-loss Aggregation employs a dynamic uncertainty-based weighting mechanism to calibrate loss contributions according to each task's uncertainty, thereby mitigating gradient interference. Evaluation on standard public benchmarks demonstrates that CLAMP consistently outperforms the vast majority of existing state-of-the-art methods.

# 1. Introduction

As social media and e-commerce platforms continue to expand, users increasingly articulate their opinions through combined image–text messages—manifesting in contexts like product reviews and news commentary. Accordingly, Multimodal Aspect-Based Sentiment Analysis (MABSA) has become a central task in affective computing [1], focused on extracting aspect terms, such as "battery life" or "screen clarity" in product evaluations, from image–text pairs and predicting their fine-grained sentiment polarities. Its research results can be widely applied in product optimization, public opinion monitoring, and personalized recommendation. Compared to traditional single-modal methods, MABSA enhances semantic understanding by fusing visual information, but it also faces challenges such as cross-modal alignment noise, task conflict, and consistency of fine-grained representation. These challenges arise from the complexity of text-image data: On one hand, sentence-level semantics often encompass multiple aspects with varying sentiment orientations, which can cause sentiment ambiguity. On the other hand, images frequently contain substantial irrelevant visual information, with only a small fraction directly pertaining to a given aspect. For instance, in the Twitter example shown in Figure 1, the text includes three distinct aspect terms, each associated with a different sentiment polarity. The emotional polarity is easily influenced by the context, such as "Cavaliers" is easily predicted as Positive due to the later text "capture NBA title" and "win". While the picture contains a lot of character information related to "Warriors", looking at the picture alone cannot obtain the emotional polarity related to "Cavaliers". This complexity requires the model to have the ability of fine-grained cross-modal alignment and dynamic noise suppression, which is not yet fully solved by existing methods.

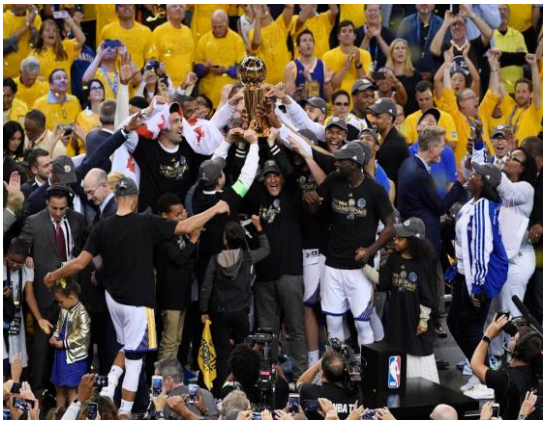

Warriors dethrone Cavaliers, capture NBA title with 129 - 120 win.

| Aspect | Cavaliers | NBA | Warriors |
|---|---|---|---|
| Sentiment | Negative | Neutral | Positive |

Figure 1 An illustrative example of MABSA task, showcasing aspects, their corresponding descriptions, and associated emotions.

Aspect-Based Sentiment Analysis (ABSA) [1] serves as a fundamental technique in social media content analysis, enabling precise extraction of aspect entities from textual data and the determination of their sentiment polarities. With the rapid evolution of digital social platforms, users are increasingly favoring multimedia formats to express their opinions and emotions. Although traditional Multimodal Sentiment Analysis (MSA) [3] methods can handle overall-level sentiment recognition tasks, they perform poorly when faced with specific aspects that require fine-grained sentiment understanding. In response, Multi-modal Aspect-based Sentiment Analysis has arisen as a formidable challenge and a prominent research focus within the field of multimodal learning. This technology combines the advantages of MSA and ABSA and enhances the extraction effect of fine-grained semantic information in text through the introduction of visual information. From the perspective of technological development, existing MABSA research can be summarized into three main directions: a pipeline processing framework, a unified modeling method based on BART [4], and cross-modal transformer technology. Within pipeline approaches, Ju et al. interpret the global alignment between text and images as the measure of visual cue integration into textual representations and introduce the Joint Multimodal Learning (JML) framework [5]. But they largely neglect object-level visual information. However, such methods rely on a sequential processing flow, which inevitably leads to the propagation of errors between subtasks. In contrast, BART can simultaneously process multimodal inputs and complete the parallel execution of multiple subtasks within the decoder, thereby mitigating error accumulation. Based on this advantage, Ling and others designed VLP-MABSA [6], a task-specific pre-training framework closely aligned with downstream MABSA tasks, which achieves end-to-end fusion of visual and textual information. However, this method still has optimization space in the intermodal alignment strategy, and only considers aligning fine-grained object visual information with text. Zhou and others proposed the AoM method [7], which focuses on uncovering semantic and emotional information tightly linked to specific aspects, but offers limited independent supervised learning for visual and textual modalities. Building on this, Yang et al. introduced the Cross-Modal Multi-Task Learning (CMMT) framework [8], incorporating a dynamic gating mechanism to regulate the influence of visual information on text processing; however, it provides only rudimentary handling of noise arising from irrelevant regions in images. Based on the CMMT framework, Xiao and others approached the problem from an aesthetic perspective, introducing the Atlantis model with aesthetic scores [9], and exploring the inherent mechanism of image emotional expression. Nevertheless, the integrated exploitation of textual syntactic features and fine-grained visual details remains suboptimal. To address this, Zou et al. proposed the Target-Oriented Cross-Modal Transformer (TCMT) [10], which comprises three modules: it harnesses syntactic information alongside optical character recognition (OCR) to extract embedded text from images and applies supervised training to the visual modality. However, the three-module structure inevitably leads to an increase in parameters.

Although these works have made some progress, there are still significant limitations: (1) Lack of alignment granularity: Most models relate graphics and texts in a global manner, ignoring the correspondence between aspect terms and local regions, resulting in irrelevant noise interference. To tackle this issue, we introduce a multi-task contrastive learning framework that aligns modality-specific information at both global and local levels. (2) Semantic coherence across modalities is hampered when image and text features are processed independently, undermining cross-modal consistency. To address this, we employ a Progressive Attention Fusion network that incrementally integrates features from both modalities throughout the model. (3) Loss optimization rigidity: In multi-task learning, the fixed weight strategy has difficulty balancing the contributions of modalities, and is easily overwhelmed by the dominant modal gradient. We added an adaptive multi-loss aggregation in the model to balance the contributions of different tasks and modalities.

To tackle these challenges, we introduce CLAMP, a Contrastive Learning framework with Adaptive Multi-loss and Progressive Attention Fusion for MABSA. CLAMP comprises four key components: a multimodal feature extraction module, a progressive attention fusion network, a multi-task contrastive learning framework, and an adaptive multi-loss aggregation module. In the feature extraction stage, two Transformer-based encoders independently derive text and visual representations. The progressive attention fusion network then employs a hierarchical attention scheme to iteratively merge and refine these representations across multiple layers, yielding a rich, unified cross-modal embedding. In the multi-task contrastive learning framework, we designed three components to cooperatively handle three tasks: the contrastive learning component, the word region alignment component, and the multi-task emotion annotation component. The contrastive learning module features a task that fosters global feature learning between text and images. Meanwhile, the word alignment module performs a fine-grained alignment task by employing the optimal transport distance to sharpen local correspondences, thereby strengthening cross-modal alignment between visual and textual features. And the multi-task emotion annotation component formulates the problem as a Conditional Random Field (CRF) model to capture label dependencies and deliver precise token-level predictions.

The principal contributions of this study are as follows:

(1) We introduce CLAMP, a fully end-to-end model, which is an innovative approach to progressively layer and fuse cross-modal information. This method does not complete the interaction between text and images all at once, but gradually deepens it through multiple stages. At each stage, the textual representations are refined using the outputs from the previous layer and then engage with the visual features through a novel cross-modal interaction. This design simulates the process of humans gradually understanding complex information, and hopes to fully integrate modal information at multiple levels and from multiple perspectives.

(2) We design a multi-task contrastive learning framework to align textual and visual representations at varying granularities, thereby reducing modality conflicts.

Moreover, we incorporate a dynamic, uncertainty-driven weighting mechanism to balance task contributions adaptively and prevent negative transfer.

(3) Extensive experiments and visualization analyses were conducted on two benchmark datasets, demonstrating that our approach consistently outperforms baseline methods and achieves superior performance on the MABSA task.

The remainder of this paper is structured as follows: Section 2 reviews related work on text-based and multimodal sentiment analysis. Section 3 details the architecture of the proposed CLAMP model. Section 4 presents and analyzes the experimental results, and Section 5 concludes the paper by summarizing the main contributions.

## 2. Related work

### 2.1 Textual aspect-based sentiment analysis

Textual aspect-based sentiment analysis (ABSA) focuses to detect fine-grained aspect terms in text and determine their corresponding sentiment polarities, making it a key research topic in affective computing. In contrast, early sentiment analysis (SA) primarily targeted coarse-grained sentiment classification at the sentence or paragraph level [11], but it is difficult to deal with the coexistence of multiple aspects of emotion in text, such as "the camera has excellent image quality but insufficient battery life", which requires separately identifying the emotional tendencies of "image quality" and "battery life". To break through this limitation, researchers propose ABSA technology [12], whose core lies in modeling the semantic association between terms and contexts. Initial approaches predominantly rely on pre-trained linguistic models, including BERT and BART, alongside neural architectures like LSTM and TextCNN, to accomplish designated tasks. Among them, LSTM captures long-distance dependencies through sequence modeling [13], while TextCNN uses local convolutional kernels to extract key phrase features [14]. In recent years, Graph Convolutional Networks (GCN) have been introduced to explicitly model syntactic dependencies, such as connecting aspect words with their modifiers through dependency trees [15]. The conventional pipeline methodology decomposes ABSA into dual subtasks: aspect term extraction and sentiment polarity classification [16], yet this approach encounters challenges related to error propagation. The end-to-end framework significantly improves robustness by jointly optimizing the subtasks through a unified model [17]. Further optimization strategies include the table filling method that converts sequence tagging into a two-dimensional matrix prediction to reduce tagging ambiguity [18], as well as multi-task learning and machine reading comprehension (MRC) techniques that enhance task synergy through parameter sharing [19] or the question and answer paradigm [20]. Despite the above methods performing excellently in single-modal text scenarios, they still have limitations such as mono-modality and insufficient fine-grained alignment: on one hand, they ignore the visual information accompanying user comments, such as product images, leading to a one-sided understanding of semantics; on the other hand, text-internal dependency modeling, such as syntax analysis, fails to combine with cross-modal interaction, making it difficult to support multi-scene applications. These shortcomings provide an important direction for the research of multi-modal sentiment analysis.

**2.2 Multimodal sentiment analysis**

Multimodal sentiment analysis (MSA), which determines emotional orientation through the fusion of diverse modal inputs including textual, visual, and auditory data, has emerged as a pivotal research area within affective computing. Current investigations primarily concentrate on conversational contexts and social media environments, where architectures including LSTM, GRU, and Transformer models have been utilized for emotional state identification [21] and irony detection tasks [22]. At the algorithmic level, multimodal fusion techniques are classified into early fusion (directly stitching features), late fusion (merging decisions after independently processing modalities), and hybrid methods (such as cross-modal consistency regression [23]). For instance, Wang et al. substantially enhanced the effectiveness of Weibo sentiment classification by integrating textual and visual characteristics via a cross-modal bag-of-words framework [24]. However, traditional MSA methods are mostly oriented towards coarse-grained sentiment classification, such as overall positive or negative judgments, which makes it difficult to meet the demand for fine-grained emotional expression in social media data. Take product reviews as an example, users may simultaneously evaluate "screen display effect" and "system smoothness", while the image may contain noise areas unrelated to specific aspects, such as background advertisements. Although some studies attempt to introduce attention mechanisms [25] or target-sensitive representations [26] to enhance modal alignment, they still lack an explicit modeling of aspect-level semantics, resulting in insufficient cross-modal noise suppression and fine-grained emotional reasoning capabilities. This limitation has spurred the development of multimodal aspect-based sentiment analysis, which seeks to identify aspects from both textual and visual modalities while determining their associated sentiment polarities, thus providing enhanced accuracy for applications including personalized recommendation frameworks and public sentiment monitoring.

**2.3 Multimodal aspect-based sentiment analysis**

MABSA combines textual and visual data to deliver more holistic sentiment analysis, garnering significant interest from the research community in recent years. In contrast to conventional text-only aspect sentiment analysis, MABSA comprises two primary subtasks: multimodal aspect sentiment classification (MASC) and multimodal aspect term extraction (MATE). MATE, formulated as a sequence-tagging problem, extracts aspect terms from text guided by visual cues; while MASC determines the emotional tendencies of these aspect terms. Recently, investigations have emerged that consolidate these dual subtasks within an end-to-end architecture, termed Joint Multimodal Aspect-Based Sentiment Analysis (JMASA) [27]. Currently, three primary challenges exist in MABSA research: cross-modal correspondence and integration; capturing inter-task dependencies; and mitigating visual interference [9]. Extraneous information within images can introduce noise, impeding the accurate extraction of aspect terms. Predominant approaches encompass attention-based mechanisms, graph neural networks, and pre-trained model architectures. Yu et al. developed an attention and fusion network emphasizing target-specific sensitivity [28], whereas Yang et al. introduced a mechanism for dynamically modulating visual information impact across various aspects [29]. The limitation of these methods is that they mostly use global attention mechanisms, ignoring the fine correspondence between aspect terms and local regions of images. Zhou et al. introduced graph convolutional networks (GCN) to learn the dependencies between two modalities, and constructed cross-modal interaction through modal aligned hidden vectors [7]. Ling et al. introduced a BART-driven generative multimodal framework for visual–language pretraining and subsequent MABSA applications [6]. Although these methods have shown significant results, they still have problems such as high computational resource requirements and reliance on a mess of pre-trained marked data. Contemporary research has also started to focus on the aesthetic properties of images, with Xiao et al. constructing a model that exploits visual aesthetic characteristics to comprehensively capture emotional expressions embedded within visual content [9]. Khan and Fu attempted to convert images into text descriptions to facilitate cross-modal alignment [30]. Although this helped alleviate the problem of modal heterogeneity, it may lead to overly neutral image descriptions and introduce new noise. Current MABSA research is moving towards fine-grained cross-modal alignment, dynamic noise suppression, and modal fusion, etc. CMMT proposes multi-aspect and emotion detection tasks for cross-modal interactive learning [8], which significantly improves the model performance. Future research trends may focus on exploring more effective visual semantic extraction and alignment techniques, designing fusion strategies that can adaptively handle different modal contributions, and developing MABSA frameworks for specific domains. As a fine-grained sentiment analysis task that integrates multimodal information, MABSA shows great potential in understanding social media content and user reviews, while also facing many challenges at the technical and application levels.

## 3. Methodology

In this section, we present the formal definition of the MABSA task, describe the overall architecture of CLAMP, and then detail the specific components that constitute the model.

### 3.1 Task formulation

Given a multimodal tweet dataset $D$, which consists of text-image pairs, the sentence text is represented as $X = (x_1, x_2, x_3, ..., x_n)$, and its corresponding image is $V_i$. The labels provided by the benchmark dataset of the MABSA task follow a unified label pattern $Y = (y_1, y_2, y_3, ..., y_n), y_i \in \{B-\text{POS}, I-POS, B-NEU, I-NEU, B-NEG, I-NEG\} \cup \{O\}$. Where $B$ denotes the beginning of an aspect term, $I$ marks tokens inside an aspect term, and $O$ indicates tokens unrelated to any aspect. $POS$, $NEU$, and $NEG$ correspond to positive, neutral, and negative sentiment polarities, respectively. Consequently, the model must perform a seven-way classification for each token $x_i$.

### 3.2 Model Overview

The architecture of the proposed CLAMP model is illustrated in Figure 2. It mainly comprises four parts: Feature Extractor, Progressive Attention Fusion network (PAF), Multi-task Contrastive Learning (MCL), and Adaptive Multi-loss Aggregation (AMA). Firstly, the Feature Extractor uses the Robustly Optimized BERT Pretraining Approach (RoBERTa) as a text encoder to process the original text, and utilizes the Vision Transformer (ViT) as an image encoder to encode the original image. Then, the encoded textual representations and visual representations are fed into PAF for feature enhancement and cross-modal feature integration. And MCL is used for further perform feature interaction between text features and image features. AMA aggregates various losses, dynamically handle various losses and assist the model in optimization, and finally outputs the predicted aspect words and their corresponding emotions.

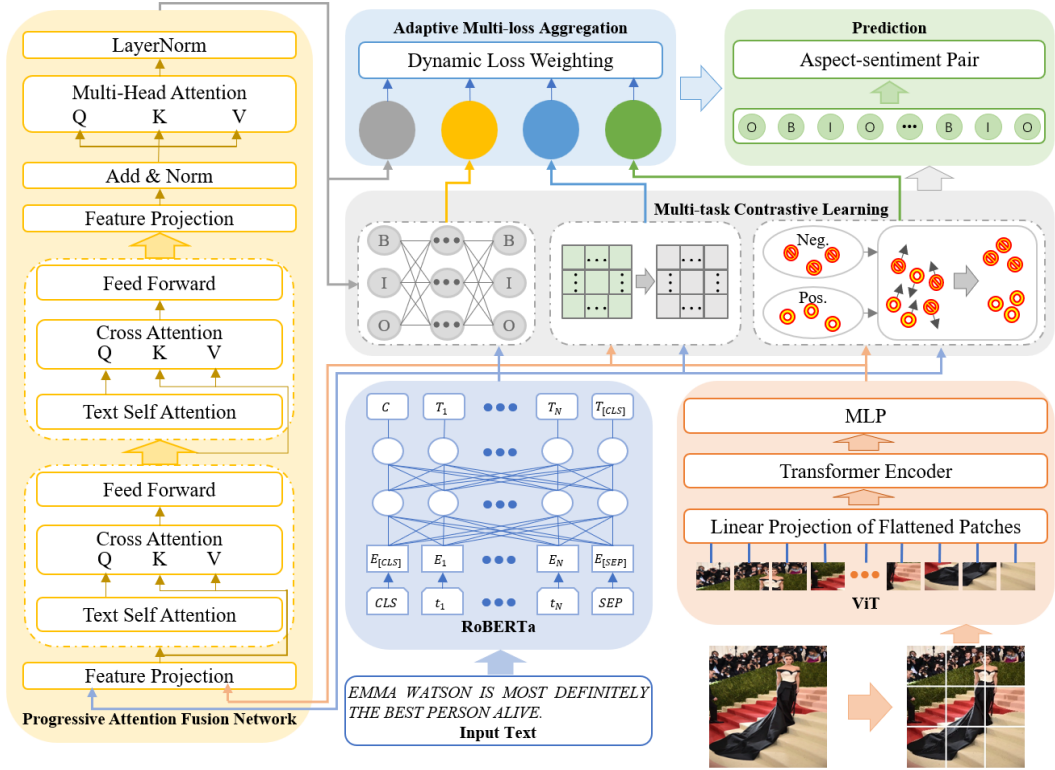

Figure 2 The overview of Contrastive Learning framework with Adaptive Multi-loss and Progressive Attention Fusion (CLAMP).

### 3.3 Feature Extractor

In the CLAMP model, the Feature Extractor serves as the foundation module of the entire architecture, responsible for extracting high-quality text and visual feature representations from the original multimodal input. Given that the MABSA task requires simultaneous processing of text sequences and image information, we employed dedicated encoders to process data from these two modalities independently, ensuring comprehensive capture of semantic information within each modality.

For text modality, we choose RoBERTa as the text encoder. RoBERTa, as an improved version of BERT, shows stronger performance in text understanding tasks by optimizing the pretraining strategy and removing the next sentence prediction task. Given an input text sequence $X = (x_1, x_2, x_3, \ldots, x_n)$, where $n$ denotes the text sequence length. The processing flow of the RoBERTa encoder includes three key steps: First, the word embedding layer transforms the token ID into $d$ dimensional embedding vector; Second, the position encoding module adds position information to each token to maintain the sequence order; Finally, the multi-layer Transformer encoder conducts deep semantic modeling via self-attention mechanisms and feedforward networks. Specifically, the RoBERTa encoder receives text features and outputs the text hidden state representation as follows:

$$H^t = \text{RoBERTa}(X) = \{h_1^t, h_2^t, \ldots, h_n^t\} \qquad (1)$$

Where $H^t \in \mathbb{R}^{n \times d}$, $d$ denotes the dimensionality of the hidden state, $h_i^t \in \mathbb{R}^d$ represents the contextual representation of the i word element. RoBERTa employs a multi-head self-attention mechanism that adeptly models long-range dependencies and intricate semantic relationships within text sequences, thereby supplying rich semantic representations for downstream aspects extraction and sentiment classification.

For image modalities, we use ViT as the visual encoder. ViT treats an image $V_i$ as a sequence by first splitting it into fixed-size patches of $P \times P$ pixels, then processing those patches through the Transformer architecture to capture both local details and global contextual information. The processing flow of ViT includes four core steps: First, the image partitioning module divides the original image into patches of fixed size; Second, the linear projection layer maps each patch to a $d$ dimensional vector space; Third, CLS classification markers and position encodings are added to maintain spatial location information; Fourth, deep visual feature learning is carried out through the Transformer encoder. The encoding process of ViT can be represented as follows:

$$H^v = \text{ViT}(V_i) = \{h^v_{[CLS]}, h^v_1, h^v_2, \ldots, h^v_k\} \qquad (2)$$

Where $H^v \in \mathbb{R}^{(k+1) \times d}$, $k$ represents the number of image patches, $h^v_{[CLS]} \in \mathbb{R}^d$ denotes the global image representation, and $h^v_j \in \mathbb{R}^d$ represents the feature representation of the jth image patch. ViT can model the spatial relationship between different image patches through the self-attention mechanism, effectively capturing visual clues related to aspect terms, while mitigating interference from irrelevant background information.

To bolster the robustness of feature representations and improve the effectiveness of model training, we perform standardization processing on the extracted text and visual features. We apply LayerNorm and Dropout regularization techniques to standardize the features, to prevent overfitting and enhance the model's generalization capability.

$$\bar{H}^t = \text{LayerNorm}\big(\text{Dropout}(H^t)\big) \qquad (3)$$
$$\bar{H}^v = \text{LayerNorm}\big(\text{Dropout}(H^v)\big) \qquad (4)$$

After the above processing, we obtained standardized text features $\bar{H}^t$ and visual features $\bar{H}^v$, which will be used as input for subsequent modules. Unlike traditional feature concatenation methods, we maintain the independence of text and visual features, and achieve more refined cross-modal interaction through a progressive attention fusion network, thus avoiding the problem of information redundancy and noise interference that may be caused by simple concatenation.

### 3.4 Progressive Attention Fusion Network

The Progressive Attention Fusion network (PAF) is the core innovative module of the CLAMP model, aiming to solve the noise interference and granularity alignment problems existing in traditional global attention mechanisms when dealing with multimodal data. Unlike existing methods that directly stitch or simply fuse cross-modal features, PAF gradually processes them through a multi-stage progressive attention mechanism. Attention modules at different stages capture different levels of cross-modal correlations. Shallow stages focus on more direct, surface correspondences e.g., color, object co-occurrence, while deep stages learn more abstract, semantic-level consistencies. Through gradual refinement, the model can more effectively fuses multimodal inputs and uncovers nuanced, complex cross-modal relationships. This usually achieves better integration results than a single, shallow interaction.

The PAF module receives standardized text features $\bar{H}^t \in \mathbb{R}^{n \times d}$ and visual features $\bar{H}^v \in \mathbb{R}^{(k+1) \times d}$ as input, where $n$ denotes the length of the text sequence, $k$ indicates the number of image blocks, and $d$ represents the feature dimension. Firstly, to ensure that different modal features can effectively interact in a unified semantic space, we project the input features into a lower-dimensional hidden space.

$$\widetilde{H}^t = \bar{H}^t W^t + b^t \quad (5)$$
$$\widetilde{H}^v = \bar{H}^v W^v + b^v \quad (6)$$

Where $W^t, W^v \in \mathbb{R}^{d \times d}$ are the projection weight matrices, and $b^t, b^v \in \mathbb{R}^{d \times d}$ are the bias vectors, $d$ is the hidden dimension. This dimensionality reduction projection not only lowers computational overhead but also aids in eliminating redundant information, thereby enhancing the compactness and efficiency of feature representations.

PAF adopts a three-stage progressive attention architecture. The first two stages use AttentionStage module, which includes three submodules: self-attention, cross-attention, and a feedforward network. This design enables the model to deepen cross-modal understanding layer by layer, from initial feature alignment to complex semantic association modeling. For the $s$ stage, $s \in \{1,2\}$, the processing flow is as follows. First, there is text self-attention enhancement, which strengthens the internal semantic association of text features through the self-attention mechanism:

$$H^{t,s} = \text{SelfATT}(\widetilde{H}^{t,s}) \quad (7)$$

Where $\widetilde{H}^{t,s}$ represents the input text features of the s stage. Subsequently, cross-attention is applied to enable interaction between textual and visual features, forming a cross-modal connection wherein the text features function as the Query and the image features serves as both Key and Value:

$$Q^s = H^{t,s} W_Q^s \quad (8)$$
$$K^s = \widetilde{H}^{v,s} W_K^s \quad (9)$$
$$V^s = \widetilde{H}^{v,s} W_V^s \quad (10)$$
$$H^s = CrossATT(Q^s, K^s, V^s) \quad (11)$$

Where $W_Q^s, W_K^s, W_V^s \in \mathbb{R}^{d \times d}$ are the projection weight matrices of the s stage. Finally, the feedforward network performs a nonlinear transformation on the features, further enhancing the representation ability. In each stage, a nonlinear transformation is performed through the feedforward network at the end:

$$H_{ffn}^s = \text{FFN}(H^s) = \text{GELU}(H^s W_1^s + b_1^s) W_2^s + b_2^s \quad (12)$$

Where $W_1^s, W_2^s \in \mathbb{R}^{d \times d}$ are the weight matrices of the s stage, and $b_1^s, b_2^s \in \mathbb{R}^{d \times d}$ are the bias vectors of the $s$ stage. To maintain consistency with the original feature dimension and facilitate subsequent module processing, we map the features back to the original dimension through an output projection layer. Finally, a residual connection is applied to integrate the original textual features with the enhanced representations, ensuring the preservation of the initial semantic information:

$$\overline{H}^{t,s} = \text{LayerNorm}(\overline{H}^t + \overline{H}^s_{ffn}) \tag{13}$$

This residual design not only prevents the problem of gradient disappearance, but also ensures that the model, while using visual information to enhance text understanding, does not lose the original text semantic content.

The feature transfer design between two progressive stages ensures the effective accumulation of information and the gradual deepening of semantic understanding. The first stage mainly focuses on basic cross-modal alignment, establishing a preliminary association between text word elements and image regions. In the second stage, building upon the first stage, it further captures more complex cross-modal semantic relationships, especially fine-grained visual clues related to aspect terms. The advantage of this gradual design is that it avoids the information loss that may be caused by dealing with complex cross-modal relationships all at once, allowing the model to learn cross-modal correspondences from shallow to deep layers in a hierarchical manner. The specialized design of each stage helps to improve overall performance. After two stages of gradual processing, we obtained text features that are fully enhanced by images. The third stage is enhanced multi-head cross-attention. Traditional multi-head attention relies on absolute position encoding, while CLAMP introduces learnable relative position biases. This configuration allows the model to more effectively encode the relative positional relationships among sequence elements:

$$P \in \mathbb{R}^{L_{max} \times L_{max}} \tag{14}$$

Where $L_{max}$ is the maximum sequence length, and $P_{i,j}$ can be learned. We add the truncated submatrix to the head attention score::

$$A_h = \frac{Q_h K_h^\top}{\sqrt{d_k}} + P \tag{15}$$

Then apply softmax row by row to obtain the attention weights, where $Q_h, K_h \in \mathbb{R}^{L_q \times d_k}$ are the query and key projections of the $h$ head, respectively. This design allows the model to automatically learn and emphasize relative positional dependencies between different attention heads. In order to balance the flow of information between new and old features, the enhanced multi-head attention uses a learnable gated residual:

$$g = \sigma(w_g) \in (0,1)^d \tag{16}$$

$$H = \text{Concat}(O_h)W_o \tag{17}$$

$$O = g \odot H + (1-g) \odot r \tag{18}$$

Where $w_g \in \mathbb{R}^d$ is the gate parameter, $O_h$ is the attention head, $r \in \mathbb{R}^{L_q \times d}$ is the sublayer input residual, and $\odot$ represents an element-wise multiplication. The $g$ obtained through Sigmoid allows different dimensions to retain the original or newly calculated features with different weights, thereby improving the flexibility and expressiveness of feature fusion. Moreover, considering that continuous normalization would weaken the information flow, we instead perform a layer normalization only once after the sublayer output:

$$\bar{O} = LayerNorm(O) \tag{19}$$

Take the text feature $H^{t,s}$ as the query, and the image feature $H^{v,s}$ as the key and value:

$$A = \text{Softmax}\left(\frac{H^{t,s}H^{v,s\top}}{\sqrt{d_k}} + P^{t,v}\right) \tag{20}$$

$$O_{t,v} = AH^{v,s} \tag{21}$$

Where $P^{t,v}$ is the relative position bias of the text image, this asymmetric attention design allows the text to prioritize the visual areas that are most relevant to the semantics of the current token when updating, thus achieving fine cross-modal alignment.

Finally, the enhanced fusion features output by PAF will also be used as input for subsequent modules, for further cross-modal contrastive learning and task optimization. Through this carefully designed progressive attention fusion network, the CLAMP model can achieve sufficient cross-modal information fusion in complex multimodal scenarios, making aspect term extraction and emotion classification more accurate.

**3.5 Multi-task Contrastive Learning**

To further refine the alignment of textual and visual representations, CLAMP model introduces a Multi-task Contrastive Learning framework (MCL). The framework consists of three submodules: global modal contrastive learning, cross-modal word region alignment, and multi-task emotion annotation. MCL receives enhanced fusion features from PAF as input, optimizes the contrast between text and image modalities at both the global and local levels, and provides rich cross-modal context information for multi-task emotion annotation. By jointly training and complementing each other, we can align text and images at different granularities, thereby improving the efficacy of text–image fusion and increases the accuracy of emotion classification in the MABSA task. In this section, we will elaborate on three sub-modules in detail.

Global contrastive learning (GCL) aims to make the global semantic representations of corresponding text-image pairs more consistent, thereby enhancing cross-modal semantic alignment. It ingests global semantic vectors from both text and images and bolsters their overall cross-modal alignment capability. Specifically, let the text encoding global feature of the $i$ sample be $x_i \in R^d$, and the image global feature be $v_i \in R^d$, both from the CLS labeled output of their respective encoders. First, we perform L2 normalization on both of them:

$$\hat{x}_i = \frac{x_i}{||x_i||_2} \tag{22}$$

$$\hat{v}_i = \frac{v_i}{||v_i||_2} \tag{23}$$

Then calculate the cosine similarity matrix $S \in R^{n \times n}$ between the text and all images, where

$$s_{ij} = sim(\hat{x}_i, \hat{v}_j) = \hat{x}_i^\top \hat{v}_j \tag{24}$$

The similarity value is scaled by the temperature parameter $\tau$ and optimized through the InfoNCE loss in the form of cross-entropy, to ensure that the correct text image pairs have the highest matching probability. The specific loss is:

$$\mathcal{L}_{\text{GCL}} = -\frac{1}{N} \sum_{i=1}^{N} \left[ \log \frac{\exp(s_{ii}/\tau)}{\sum_{j=1}^{N} \exp(s_{ij}/\tau)} + \log \frac{\exp(s_{ii}/\tau)}{\sum_{j=1}^{N} \exp(s_{ji}/\tau)} \right] \tag{25}$$

The first and second items respectively correspond to the two-way comparison of text or images as queries, and $N$ is the batch size. The contrast loss $\mathcal{L}_{\text{GCL}}$ outputs a scalar, which is used to enhance the alignment of cross-modal global semantics. This module optimizes similarity by pulling together representations of semantically matching text–image pairs and pushing apart those of non-matching pairs, thereby enhancing the semantic coherence of their subsequent fusion.

Unlike global contrast learning, which focuses on overall semantic consistency, the word region alignment module (WRA) focuses on local fine-grained alignment. The module receives as inputs the normalized textual token features and the image patch representations. Specifically, let the text sequence contain $n$ word elements, whose encoded features are $x_1, x_2, \ldots, x_n$, and the image is divided into $k$ patches, whose features are $p_1, p_2, \ldots, p_k$, all of which are $d$ dimensional vectors and have been L2 normalized. First, calculate the cosine distance matrix $C \in R^{n \times k}$ between all word elements and all image regions, where

$$C_{ij} = 1 - \cos(x_i, p_j) = 1 - \frac{x_i^\top p_j}{||x_i||_2 \, ||p_j||_2} \tag{26}$$

To obtain a one-to-one matching relationship between word element regions, we introduce the Iterative Optimal Transport (IPOT) algorithm. This algorithm minimizes the weighted distance sum by iteratively updating the optimal transport matrix $T \in R^{n \times k}$. During each iteration, the matching probabilities are revised using the current transport and distance matrices, ultimately producing an approximate sparse transport matrix. In actual implementation, we only take the optimal matching region of each word element to obtain the mapping relationship of the word region. Finally, we accumulate the distance values of all matching pairs to form the alignment loss:

$$\mathcal{L}_{WRA} = \sum_{i=1}^{n} \sum_{j=1}^{k} T_{ij} \, C_{ij} \tag{27}$$

where $T_{ij} \in {0,1}$ represents whether the word element $t_i$ matches the local area $p_j$. $\mathcal{L}_{WRA}$ measures the representation difference between words and local areas of images. By minimizing this loss, the model learns to maintain the consistency of cross-modal representation at the fine-grained level. This is particularly important for the MABSA task, as the target entity or emotional clues often only appear in local areas of images, and local alignment can help the model capture this detailed information.

The multi-task emotion labeling task module is specifically designed for multi-modal emotion analysis. We set up two independent sequence classifiers: the first classifier takes the fused multi-modal features as input, and the second classifier takes the original text features as input. Specifically, suppose that after cross-modal fusion, we obtain the joint representation of text and image as $h_{PAM} \in R^{n \times d}$, and the output of the original text encoding is $h_t \in R^{n \times d}$. After the classifier performs a linear transformation on $h_{PAM}$, it obtains the seven category sentiment label scores for each word element, $u_i^{(0)} \in R^7$; similarly, it obtains $u_i^{(1)} \in R^7$ after performing a linear transformation on $h_t$. To capture the correlation between label sequences, the output of each classifier is decoded in the CRF (Conditional Random Field) layer. The model uses two sets of CRF structures to calculate the negative log-likelihood loss of the label sequence, which are denoted as $\mathcal{L}_{CRF0}$ and $\mathcal{L}_{CRF1}$. Between the two classifiers, we introduce a residual connection to enhance information flow: that is, we add the original text features $h_t$ to the PAF output, forming a mixed feature input to the classifier. Specifically, it can be expressed as:

$$\tilde{h}_{PAM} = h_{PAF} + W_r h_t \tag{28}$$

where $W_r$ is a linear transformation matrix that matches the feature dimension. The residual connection helps the gradient to be directly passed back from the output layer to the original text encoding, improving the training stability. The final classification and sequence tagging total loss is the sum of two CRF losses:

$$\mathcal{L}_{CRF} = \mathcal{L}_{CRF0} + \mathcal{L}_{CRF1} \tag{29}$$

This module performs emotion classification from two complementary perspectives: cross-modal fusion features and unimodal textual features. This two-pronged strategy allows the model to leverage rich multimodal context while maintaining the inherent semantics of the textual input, thereby enhancing both the accuracy and robustness of sentiment classification in the MABSA task.

The above three submodules work together and promote each other. The global contrastive learning module provides cross-modal similarity constraints at the semantic level, roughly aligning text and images; the word region alignment module aligns fine-grained information at the local level, making the detailed representations between modalities more consistent; the multi-task emotion annotation module improves the final emotion prediction performance through supervised signals, and combines global and local information through residual connections.

## 3.6 Adaptive Multi-loss Aggregation

In the CLAMP model, the training process includes multiple tasks, each of which corresponds to a loss term. Specifically, these include the text classification loss $\mathcal{L}_{CLS}$, the multi-task emotion labeling loss $\mathcal{L}_{CRF}$, the global modal contrast loss $\mathcal{L}_{GCL}$, and the word region alignment loss $\mathcal{L}_{WRA}$.

To effectively coordinate these losses, we introduce a dynamic uncertainty weighting strategy, combining the ideas of uncertainty weighting, dynamic task priority, and fixed weights, to dynamically weight each task loss. Specifically, for each task $i$, total $M = 4$ tasks, we introduce a learnable uncertainty parameter $\sigma_i > 0$ and a task priority parameter $\pi_i$, and define a smoothing parameter $\alpha \in [0,1]$ and a temperature parameter $\tau > 0$. First, we calculate the normalized weight of task priority:

$$\omega_i = \frac{\exp(\pi_i/\tau)}{\sum_{j=1}^{M} \exp(\pi_j/\tau)} \quad (30)$$

Where the larger $\pi_i$ indicates that task $i$ currently has a higher priority, for instance, the model demonstrates comparatively low performance on this task. Then, we integrate fixed uniform weights and dynamic weights:

$$\widehat{\omega}_i = (1-\alpha)\frac{1}{M} + \alpha\omega_i = (1-\alpha)\frac{1}{M} + \alpha\frac{\exp(\pi_i/\tau)}{\sum_j \exp(\pi_j/\tau)} \quad (31)$$

Based on this, we introduce the contribution of uncertainty weighting and obtain the weighted items for each task. The final total loss is in the form of:

$$\mathcal{L} = \sum_{i=1}^{M} \widehat{\omega}_i \left(\frac{1}{2\sigma_i^2}\mathcal{L}_i + \log\sigma_i\right) \quad (32)$$

where $\mathcal{L}_i$ represents $\mathcal{L}_{CRF}$, $\mathcal{L}_{CLS}$, $\mathcal{L}_{GCL}$, and $\mathcal{L}_{WRA}$ respectively. Here, the factors before each loss, $\frac{1}{2\sigma_i^2}$ and $\log\sigma_i$, come from the theory of uncertainty weighting. Their function is: the larger $\sigma_i$ is, indicating the higher uncertainty of the task, the lower the weight of the corresponding loss term; conversely, the smaller $\sigma_i$ is, the loss term is amplified. At the same time, $\log\sigma_i$ plays a regularization role, preventing $\sigma_i$ from increasing infinitely. The parameter $\alpha$ is used to smoothly transition from the uniform weight $\frac{1}{M}$ to the dynamic priority $\omega_i$ during training, while the temperature $\tau$ controls the "sharpness" of the priority distribution. A smaller $\tau$ will make $\omega_i$ more focused on the task with the maximum $\pi_i$. This weighting strategy combines the fixed proportion commonly used in multi-task learning, the adaptive weights based on uncertainty, and the dynamic priority based on difficulty, ensuring that the training process pays appropriate attention to all tasks without being overly dominated by a single task.

# 4. Experiments

## 4.1 Experimental settings

**Datasets:** We assess the performance of our proposed approach on two widely adopted benchmark datasets: Twitter2015 and Twitter2017. Zhang et al. originally provided these two datasets [31]. Lu et al. labeled the aspect sentiment polarity [32], and Ling et al. revised the dataset [6]. Table 1 provides an overview of the key properties of the datasets., with both containing paired textual and visual information. The text content includes the corresponding image file names, aspect terms, sentences, and emotion labels, where 0, 1, 2 respectively represent negative, neutral, and positive emotional polarities. The specific aspects and emotional information are provided in Table 2.

Table 1 Statistics of the benchmark datasets. (AL: Average Length, ML: Max Length)

| Datasets | Twitter-2015 | | | Twitter-2017 | | |
| --- | --- | --- | --- | --- | --- | --- |
| | Train | Dev | Test | Train | Dev | Test |
| Positive | 928 | 303 | 317 | 1508 | 515 | 493 |
| Neutral | 1883 | 670 | 607 | 1638 | 517 | 573 |
| Negative | 368 | 149 | 113 | 416 | 144 | 168 |
| Total | 3179 | 1122 | 1037 | 3562 | 1176 | 1234 |
| Image | 3179 | 1122 | 1037 | 3562 | 1176 | 1234 |
| Sentence | 2101 | 727 | 674 | 1746 | 577 | 587 |
| AL | 16.7 | 16.7 | 17.0 | 16.2 | 16.4 | 16.4 |
| ML | 35 | 40 | 37 | 39 | 31 | 38 |

Table 2 Descriptive statistics of the two benchmark datasets. (OA: One Aspect, MA: Multiple Aspects, MS: Multiple Sentiments)

| Datasets | OA | MA | MS | Sentence |
| --- | --- | --- | --- | --- |
| Twitter-2015 | 2159 | 1343 | 1257 | 3502 |
| Twitter-2017 | 976 | 1934 | 1690 | 2910 |

**Implementation details:** For the textual modality, we utilize the RoBERTa model to capture grammatical and contextual features, while for the visual modality, the ViT model is employed to extract informative representations from images. The initial settings of specific hyperparameters for the experiment are shown in Table 3.

Table 3 Specific parameter Settings.

| Hyperparameters | Initial Setting |
| --- | --- |
| Roberta dim | 768 |
| Vit dim | 768 |
| Patch size | 16×16 |
| Epochs | 50 |
| Batch size | 32 |
| Learning rate | 2e-5 |
| L2 regularization | 0.01 |
| Optimizer | Adamw |
| Dropout rates | 0.5 |
| Self attention heads | 12 |
| Cross attention heads | 12 |

**Evaluation Metrics:** We employ precision, recall, and the micro-averaged F1 score as the evaluation criteria for MABSA. A prediction is considered accurate only if it correctly identifies both the aspect term and its associated sentiment polarity.

**4.2 Baseline models**

**4.2.1  Textual approaches**

SPAN [33] is a hierarchical, span-based end-to-end approach for ABSA. It utilizes an LSTM-based multi-span decoding mechanism to extract aspect terms and subsequently predicts sentiment based on the representations of the identified spans.

GPT-2 [34] realizes the end-to-end application of ABSA via text generation. It adopts the Transformer architecture and only uses its decoder structure.

D-GCN [35] is a Directional Graph Convolutional Network based on BERT. It captures the correlation between words through sequence tagging. It integrates syntactic dependencies to simultaneously detect aspects and their emotional polarities.

RoBERTa [36] is an improved method based on BERT. It inputs the text representation into the Transformer encoder, then uses the CRF to complete the sequence tagging.

BART [37] reformulates the end-to-end ABSA task as an index generation problem, allowing the pre-trained BART model to effectively handle each subtask within the unified framework.

**4.2.2  Multimodal approaches**

UMT+TomBERT and OSCGA+TomBERT [38] are two pipeline-based approaches that leverage UMT [39] or OSCGA [40] to fuse textual and visual features for aspect identification. These methods then utilize the target-oriented multimodal

TomBERT to predict the sentiment polarity of the identified aspects.

UMT-collapse [39]、OSCGA-collapse [40] and RpBERT-collapse [41] all use collapsed labels to represent aspect and sentiment pairs, and they are all based on the method of multi-modal aspect target extraction (MATE). UMT-collapsed improves UMT by using collapsed labels to complete the MABSA task. OSCGA-collapsed uses collapsed labels to improve OSCGA. RpBERT-collapsed adopts a multi-task learning framework to perform image-text relationship detection, integrating collapsed labels to enhance model performance and robustness.

CLIP [42] is a visual-language pre-training model that leverages contrastive learning to generate rich semantic embeddings of both text and images, enabling effective modeling of multimodal inputs.

JML [43] employs a multi-task learning framework to address the MATE and MASC subtasks. It employs a hierarchical architecture that incorporates a visual gating mechanism into the Transformer layers, enhancing the model's ability to process and integrate multimodal information.

CapTrRoBERTa [44] utilizes the DETR to transform visual inputs into textual descriptions. These generated captions are then concatenated with the original text and input into the RoBERTa model for subsequent processing.

VLP-MABSA [6] is a visual-language pre-training approach tailored specifically for the MABSA. This method leverages the BART model to fuse multimodal information and jointly model aspects, opinions, and their consistency within multimedia contexts. It employs five specialized pre-training tasks designed to simulate aspects, opinions, and cross-modal alignment.

CMMT [8] is a multi-task learning framework that integrates two auxiliary tasks aimed at steering the generation of intra-modal representations. Furthermore, it introduces a multimodal gating mechanism to dynamically modulate the influence of visual information during cross-modal interaction modeling.

GMP [45] automatically creates aspect- and emotion-focused prompts in text-image scenarios with limited data, facilitating multimodal emotion analysis.

M2DF [46] improves upon VLP-MABSA by incorporating coarse-grained and fine-grained noise metrics to assess the level of noise in training images. It adopts a twofold strategy to effectively mitigate the negative impact of image noise.

MOCOLNet [47] is a momentum contrast learning network that integrates the pre-training and training phases into a unified end-to-end framework. It requires fewer labeled samples specifically for sentiment analysis while achieving improved prediction performance. The model incorporates a multimodal contrastive learning approach alongside an auxiliary momentum strategy to enhance robustness.

JCC [48] is a joint modal cyclic complementary attention framework that improves the model's understanding of aspect relevance by combining global text and optimizing aspect extraction and sentiment classification. It uses text for visual highlighting to reduce the impact of visual noise. And it designs a cyclic attention module for aspect extraction focusing on general features and a modal complementary attention module for sentiment classification focusing on detailed information.

DualDe [49] consists of two distinct components: a hybrid curriculum denoising component that enhances sentence-image denoising through adaptive curriculum learning strategies, and an aspect enhancement denoising module that reduces aspect-related noise using an aspect-guided attention mechanism. Together, these components address challenges related to both sentence-image and aspect noise.

MCPL [50] employs a multi-model collaborative guided progressive learning approach, exploiting correlations between task-specific models and downstream tasks to enlarge high-quality training datasets. It provides progressive supervision signals that enhance model adaptability by progressively advancing from simpler to more challenging tasks.

**4.3 Main results**

Table 4 presents a performance comparison of the proposed CLAMP model against various baseline approaches on the Twitter benchmark datasets [51]. As indicated, CLAMP attained an F1 score of 67.7% on Twitter-2015 and 68.9% on Twitter-2017, showcasing its strong capabilities in multimodal sentiment analysis.

First of all, in the text-based methods, BART and RoBERTa performed excellently in the single-modal baseline, confirming the effectiveness of large-scale pre-trained language models for text representation [52]. As an encoder-decoder architecture, the BART model excels in text representation and sequence generation, achieving F1 scores of 63.9% and 65.4% on the Twitter datasets, respectively. RoBERTa, as an improved version of BERT, demonstrated robust results in text field, attaining an F1 score of 66.2% on Twitter-2017, surpassing BART, and also achieving an F1 value of 63.5% on Twitter-2015. However, these single-modal models lack the utilization of visual information, limiting their performance in multimodal scenarios. Compared with BART, our CLAMP model increased the F1 score by 3.8% and 3.5% severally in two datasets, and compared with RoBERTa, CLAMP enhanced the F1 score by 4.2% and 2.7% in two datasets, fully proving the important value of multimodal information fusion for aspect sentiment analysis.

Secondly, in the multimodal approach, the joint approach notably outperforms the collapsed method. The collapsed strategy divides the MABSA task into two independent subtasks, which may lead to error propagation [53]. In contrast, the unified labeling strategy employed by the Joint method mitigates this issue by avoiding such propagation. Similarly, BART-based methods like VLP-MABSA and M2DF effectively prevent error propagation across subtasks. Among them, the CMMT, M2DF, and MCPL models show excellent performance. The F1 values of CMMT on two datasets are 66.5% and 68.5%, respectively. Compared to the CMMT model, CLAMP achieves improvements of 1.2% and 0.4%, respectively. While CMMT generates intra-modal representations through auxiliary tasks and utilizes a multimodal gating mechanism to regulate visual information contribution, it falls short in effectively handling irrelevant regional noise present in images. M2DF is an improvement based on the VLP-MABSA method, and its F1 scores are improved to 67.6% and 68.3% respectively. In comparison, the CLAMP model improved F1 scores by 0.1% and 0.6% in two datasets. M2DF effectively mitigates the negative effects of image noise by introducing both coarse-grained and fine-grained noise metrics to quantify noise [54], employing a twofold strategy. MCPL got F1 scores of 67.6% and 68.1% on the datasets, with CLAMP further enhancing these results by 0.1% and 0.8%. MCPL leverages correlations between task-specific models and downstream tasks to augment high-quality training datasets and delivers progressive supervision signals to improve model adaptability [50]. However, the inclusion of teacher models increases training complexity and costs, and it does not adequately capture visual emotional cues or filter visual noise.

Finally, our proposed CLAMP model shows excellent performance. CLAMP model fails to achieve the optimal recall rate on the Twitter-2015 dataset, but it surpasses GMP and MCPL in other metrics of both datasets. The superior performance of the CLAMP over other baseline methods can be attributed to the following factors: (1) Fully integrating the fine-grained features of text and images. CLAMP fully leverages fine-grained information from both text and images via a multi-stage attention fusion mechanism, and more refined information flow control can more effectively capture the complex relationship between textual and visual features. This is underexplored in the CMMT, M2DF, and MCPL models [55]. Effective fusion of modal information is essential for precise sentiment identification of aspect terms. (2) We employ a multi-task contrastive learning framework to capture semantic and structural relationships between the two modalities from multiple perspectives, thereby constructing a more robust cross-modal representation. By using contrastive learning, we map semantically related texts and images to similar representation spaces, capturing the corresponding relationship globally, while word region alignment focuses on local fine-grained alignment, and uses sequence tagging to analyze the structural relationship between modalities. (3) Adaptive multi-task balancing. Dynamically adjust the weights of the losses of each task to avoid competition and interference between tasks, while minimizing the impact of noise interference during training. Automatically adjusting the task priority according to the learning progress allows the model to better extract complex modal features and integrates them, thereby mitigating conflicts between modalities.

Table 4 Results (%) of different methods for MABSA in two Twitter datasets. The best results are marked in cyan, and the second-best results are marked in bright green.

| Modality | Methods (Venue) | Twitter-2015 | | | Twitter-2017 | | |
|---|---|---|---|---|---|---|---|
| | | P | R | F | P | R | F |
| Text-based | SPAN (ACL2019) | 53.7 | 53.9 | 53.8 | 59.6 | 61.7 | 60.6 |
| | GPT-2 (2019) | 66.6 | 60.9 | 63.6 | 55.3 | 59.6 | 57.4 |
| | D-GCN (COLING2020) | 58.3 | 58.8 | 59.4 | 64.2 | 64.1 | 64.1 |
| | BART (ACL2021) | 62.9 | 65.0 | 63.9 | 65.2 | 65.6 | 65.4 |
| | RoBERTa (IP&M2022) | 61.8 | 65.3 | 63.5 | 65.5 | 66.9 | 66.2 |
| Multimodal | UMT+TomBERT (ACL2020) | 58.4 | 61.3 | 59.8 | 62.3 | 62.4 | 62.4 |
| | OSCGA+TomBERT (MM2020) | 61.7 | 63.4 | 62.5 | 63.4 | 64.0 | 63.7 |
| | UMT-collapsed (ACL2020) | 60.4 | 61.6 | 61.0 | 60.0 | 61.7 | 60.8 |
| | OSCGA-collapsed (MM2020) | 63.1 | 63.7 | 63.2 | 63.5 | 63.5 | 63.5 |
| | RpBERT-collapsed (AAAI2021) | 49.3 | 46.9 | 48.0 | 57.0 | 55.4 | 56.2 |
| | CLIP (ICML2021) | 44.9 | 47.1 | 45.9 | 51.8 | 54.2 | 53.0 |
| | JML (EMNLP2021) | 65.0 | 63.2 | 64.1 | 66.5 | 65.5 | 66.0 |
| | CapTrRoBERTa (MM2021) | 60.6 | 66.1 | 63.2 | 67.1 | 67.4 | 67.3 |
| | VLP-MABSA (ACL2022) | 65.1 | 68.3 | 66.6 | 66.9 | 69.2 | 68.0 |
| | CMMT (IPM2022) | 64.6 | 68.7 | 66.5 | 67.6 | 69.4 | 68.5 |
| | GMP (ACL2023) | 65.5 | 68.8 | 67.1 | 66.8 | 68.0 | 67.4 |
| | M2DF (EMNLP2023) | 67.0 | 68.3 | 67.6 | 67.9 | 68.8 | 68.3 |
| | MOCOLNet (TKDE2023) | 66.3 | 67.9 | 67.1 | 67.3 | 68.7 | 68.0 |
| | JCC (ICMEW2024) | 63.3 | 63.4 | 63.3 | 67.3 | 65.2 | 66.2 |
| | DualDe (PACLIC2024) | 66.1 | 68.2 | 67.1 | 66.4 | 68.2 | 67.3 |
| | MCPL (KBS2024) | 66.4 | 68.9 | 67.6 | 67.2 | 69.0 | 68.1 |
| Ours | CLAMP | 67.2 | 68.3 | 67.7 | 68.2 | 69.7 | 68.9 |

### 4.4 Ablation study

Table 5 Ablation study results (%) for the CLAMP model. The best performances are highlighted in cyan.

| Methods | Twitter-2015 | | | Twitter-2017 | | |
|---|---|---|---|---|---|---|
| | P | R | F | P | R | F |
| CLAMP | 67.2 | 68.3 | 67.7 | 68.2 | 69.7 | 68.9 |
| w/o PAF | 63.2 | 64.8 | 64.0 | 66.0 | 67.8 | 66.9 |
| w/o MCL | 63.5 | 67.1 | 65.3 | 67.1 | 67.4 | 67.3 |
| w/o AMA | 62.4 | 65.4 | 63.9 | 67.0 | 68.3 | 67.7 |

We performed multiple ablation experiments to assess the impact of each component within CLAMP, with the results are presented in Table 5 and Figure 3. Specifically, we studied the impact of the following components: (1) "w/o PAF" represents removing the progressive attention mechanism from CLAMP. (2) "w/o MCL" represents removing the multi-task contrastive learning framework from the model. (3) "w/o AMA" means to remove the adaptive multi-loss aggregation in the framework and use a simple linear summation method, $\mathcal{L} = \mathcal{L}_{CRF} + \mathcal{L}_{CLS} + \mathcal{L}_{GCL} + \mathcal{L}_{WRA}$. The experimental results indicate that all three components substantially influence the model's results, with the progressive attention mechanism exerting the greatest effect, followed by the adaptive multi-loss aggregation module, and lastly the multi-task contrastive learning framework.

The outcomes of the ablation study can be explained by several key factors. Primarily, the progressive attention fusion network exerts the most significant influence on model's performance, as its progressive fusion strategy enables CLAMP to more effectively capture and integrate rich multimodal information. By gradually enhancing feature interaction through multi-stage attention processing, each stage captures different levels of cross-modal relationships, forming a hierarchical feature representation. Secondly, the model cannot perceive multimodal information from multiple angles without a multi-task contrast learning framework, and its removal also has a significant impact. Finally, the adaptive multi-loss aggregation can fully utilize the interactive information from the progressive attention fusion module and the multi-task contrastive learning framework. If it is removed, the model cannot fully utilize the modal information. As a result, when the adaptive multi-loss aggregation is removed from the CLAMP framework, the model's performance drops significantly.

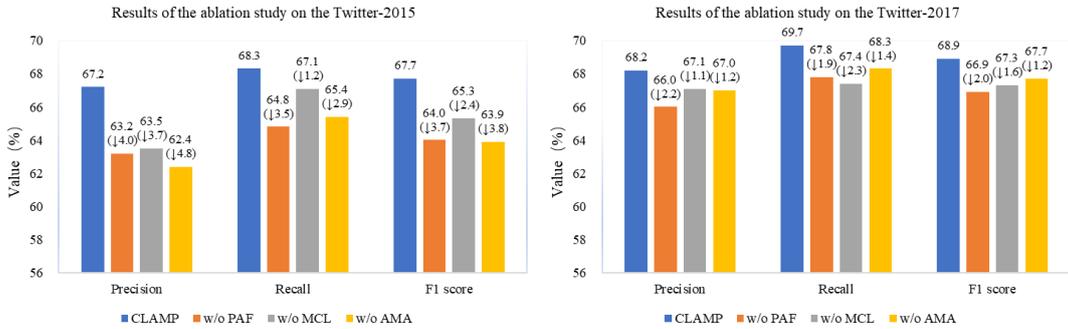

Figure 3 Comparison of ablation study results.

## 4.5 Case study

To validate CLAMP's performance on Twitter comments, we conducted a case study comparing it against two baseline models with publicly available code: VLP-MABSA and MCPL. The experimental outcomes are depicted in Figure 4. In the cases presented in Figure 4(a), the text includes multiple aspect terms, while the images feature complex and diverse backgrounds, significantly challenging the baseline models' capacity to accurately extract aspect terms. The VLP-MABSA model cannot accurately identify the aspect term "mayor". MCPL model can accurately predict the corresponding sentiment polarity, but it does not extract the complete aspect term "Mayor Kadokawa", only extracting part of the aspect terms. In the example shown in Figure 4(b), the two aspect terms contained in the sample correspond to opposite emotional polarities, and include complex human image information. The VLP-MABSA model can accurately extract the aspect term "Blackhawks" but cannot accurately predict its corresponding emotional polarity. In the example presented in Figure 4(c), the image offers accurate and relevant information that supports the MABSA task, but the VPL-MABSA model may be influenced by the word "Seriously" before the aspect term "Colgate", predicting the emotional polarity as Negative instead of the correct Neutral. While the MCPL model mistakenly considered "toothpaste" as an aspect term, failing to fully leverage the visual information. The CLAMP model can accurately identify aspect terms in three case samples and correctly predict the corresponding emotions. The results of the case study show that CLAMP effectively captures critical information from both images and text through progressive multimodal fusion and multi-task contrastive learning, which can effectively handle complex and diverse multi-modal data, thereby improving the performance of the MABSA task.

| | | | |
|---|---|---|---|
| Image | 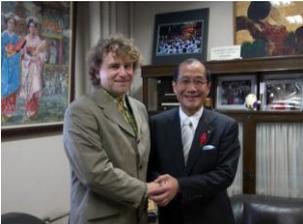 | 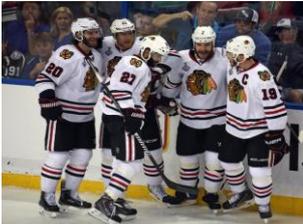 | 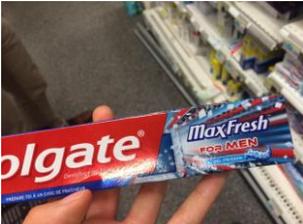 |
| Text | (a) Complain about Kyoto a bit and someone takes you to see the mayor. Interesting! Mayor Kadokawa, thanks for your time! | (b) Brent Seabrook celebrates his goal, but the good times didn't last. #Blackhawks fall. | (c) RT @ kjalee: Seriously Colgate, do we really need gendered toothpaste? |
| VLP-MABSA | (Kyoto, Negative 😒) (-)✗ (Mayor Kadokawa, Positive 😀) | (Brent Seabrook, Positive 😀) (Blackhawks, Positive 😀)✗ | (Colgate, Negative 😒)✗ |
| MCPL | (Kyoto, Negative 😒) (mayor, Neutral 😐) (Kadokawa, Positive 😀)✗ | (Brent Seabrook, Positive 😀) (Blackhawks, Negative 😒) | (toothpaste, Neutral 😐)✗ |
| CLAMP | (Kyoto, Negative 😒) (mayor, Neutral 😐) (Mayor Kadokawa, Positive 😀) | (Brent Seabrook, Positive 😀) (Blackhawks, Negative 😒) | (Colgate, Neutral 😐) |

Figure 4 Case study results. Aspects related to positive emotions are marked in yellow, aspects related to neutral emotions are marked in cyan, and aspects related to negative emotions are marked in bright green.

# 5. Conclusion

The paper presents an end-to-end Contrastive Learning framework with Adaptive Multi-loss and Progressive Attention Fusion (CLAMP) tailored for the MABSA task. CLAMP delves into deeper semantic and structural features across modalities, enabling dynamic alignment, cross-modal interaction, and efficient information fusion. CLAMP consists of three main modules: the Progressive Attention Fusion network (PAF), Multi-task Contrastive Learning (MCL), and Adaptive Multi-loss Aggregation (AMA). Among them, PAF achieves deeper and more refined cross-modal information fusion through a progressive interaction method. MCL uses multi-task contrast to learn cross-modal knowledge from different angles and granularities, while AMA uses a dynamic uncertainty weighting strategy to intelligently coordinate the learning process of each task, avoiding negative transfer and task conflicts between tasks. Compared to prior approaches, CLAMP demonstrated superior results on two benchmark datasets, highlighting that capturing deeper structural and semantic features enhances the alignment and integration of multimodal information. Furthermore, ablation studies confirmed the critical roles of the progressive fusion strategy, multi-task contrastive learning framework, and Adaptive Multi-loss Aggregation, which function synergistically to preserve information completeness. Finally, the case study experiment showed that CLAMP can effectively handle complex and diverse modal information. However, our work still has some shortcomings, such as the adequacy of progressive fusion being influenced by the feature extraction model, and the impact of multi-loss aggregation plays a pivotal role in determining the model's overall performance. We believe that using progressive hierarchical interaction and multi-task contrast to overcome the modal gap will stimulate the interest and creativity of more researchers.

## Declaration of generative AI and AI-assisted technologies in the writing process

During the preparation of this work the authors used Claude and ChatGPT in order to improve readability and language of the work. After using these tools, the authors reviewed and edited the content as needed and take full responsibility for the content of the published article.

## Data availability

Data will be made available on request.